\documentclass[10pt,twocolumn,letterpaper]{article}

\usepackage{mpi}
\usepackage{times}
\usepackage{epsfig}
\usepackage{graphicx}
\usepackage{amsmath}
\DeclareMathOperator*{\argmax}{arg\,max}
\usepackage{amssymb}
\usepackage{multirow}
\usepackage{booktabs}
\usepackage{mathtools}

\usepackage[labelsep=period,labelfont=bf, font=small]{caption}
\usepackage{algpseudocode,algorithm,algorithmicx}
\usepackage{xr}
\usepackage{bm}
\usepackage{color}
\usepackage[hyphens]{url}

\usepackage[pagebackref=true,breaklinks=true,colorlinks,bookmarks=false]{hyperref}

\mpifinalcopy

\begin{document}

\title{High-Fidelity Neural Human Motion Transfer from Monocular Video\vspace{-7pt}} %

\author{
\hspace{1.6em}Moritz Kappel$^1$\hspace{1.8em}
Vladislav Golyanik$^2$\hspace{1.8em}
Mohamed Elgharib$^2$\hspace{1.5em}
Jann-Ole Henningson$^1$\vspace{0.3em}\\
Hans-Peter Seidel$^2$\hspace{2.1em}
Susana Castillo$^1$\hspace{2.6em}
Christian Theobalt$^2$\hspace{3.0em}
Marcus Magnor$^1$\vspace{0.5em}\\
$^{1}$ICG, TU Braunschweig\hspace{2.4em}
$^{2}$MPI for Informatics, SIC 
}

\twocolumn[{ 
\renewcommand\twocolumn[1][]{#1} 
\maketitle 
\begin{center} 
    \includegraphics[width=1.0\textwidth]{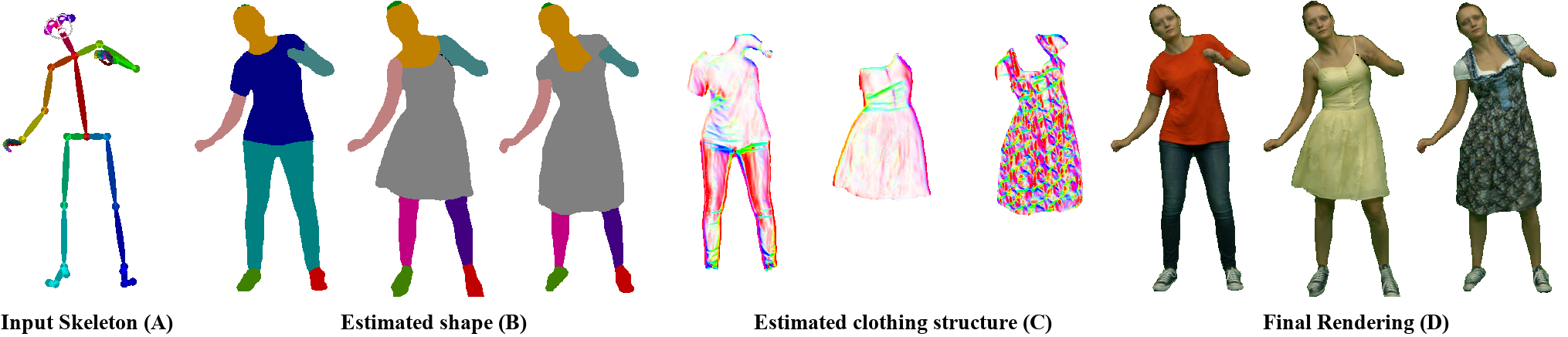} 
    \captionof{figure}{
    \textbf{Our human motion transfer framework} translates 2D pose skeletons (\textbf{A}) to photorealistic images (\textbf{D}) by explicitly estimating the actor's 2D shape (\textbf{B}) and internal structure of clothing (\textbf{C}), enabling direct deformation handling and support of high-frequency details such as wrinkles and shadows.} 
    \label{fig:teaser} 
\end{center} 
}] 

\begin{abstract} 
\vspace{-4pt} 
Video-based human motion transfer creates video animations of humans following a source motion. Current methods show remarkable results for tightly-clad subjects.
However, the lack of temporally consistent handling of plausible clothing dynamics, including fine and high-frequency details, significantly limits the attainable visual quality. 
We address these limitations for the first time in the literature and present a new framework which performs high-fidelity and  temporally-consistent human motion transfer with natural pose-dependent non-rigid deformations, for several types of loose garments. 
In contrast to the previous techniques, we perform image generation in three subsequent stages, synthesizing human shape, structure, and appearance. 
Given a monocular RGB video of an actor, we train a stack of recurrent deep neural networks that generate these intermediate representations from 2D poses and their temporal derivatives.
Splitting the difficult motion transfer problem into subtasks that are aware of the temporal motion context helps us to synthesize results with plausible dynamics and pose-dependent detail. It also allows artistic control of results by manipulation of individual framework stages. 
In the experimental results, we significantly outperform the state-of-the-art in terms of video realism. 
Our code and data will be made publicly available\footnote{\url{https://graphics.tu-bs.de/publications/kappel2020high-fidelity}}. 
\end{abstract} 
\vspace{-11pt} 
\section{Introduction}\label{sec:intro}
Human motion transfer methods, also known as performance cloning or reenactment methods, can generate realistic video animations of an actor following a target motion specified by a user. 
This has several applications in AR/VR and video editing. 
Building upon new advances in machine learning, current motion transfer methods tackle this challenging problem by learning a direct mapping between an actor-independent motion space and the resulting target actor's appearance space. 
These methods often require a training video of an actor performing a rich set of motions~\cite{Chan2019,Shysheya2019,lwb2019,liu2020neural,bansal2018recycle,wang2018pix2pixHD}. 

Some recent motion transfer approaches parameterize motion as skeletal pose sequences that can be computed from videos with off-the-shelf pose detectors~\cite{Chan2019,Shysheya2019}. 
Others use pre-captured template meshes or parameterized body models to provide additional guidance to the synthesis step~\cite{lwb2019,liu2020neural}. 
Acquisition of such templates~\cite{liu2020neural}, however, requires an extensive structure-from-motion reconstruction of the static target actor under constant lighting. 
Furthermore, existing human motion transfer approaches are likely to produce 
notable temporal and spatial artifacts when actors wear loose clothing, such as dresses, skirts and hoodies~\cite{Chan2019,bansal2018recycle,wang2018pix2pixHD,lwb2019}. 
On such garments, they struggle to realistically reproduce the appearance of fine-scale details like folds and wrinkles, as well as plausible dynamics. 

In this paper, we present a new human motion transfer framework that generates visually plausible video animations of humans that are spatially and temporally coherent, and show natural dynamics, even for actors wearing loose garments (see Fig.~\ref{fig:teaser}). 
Given a single monocular video of an actor performing a rich set of motions, we train a stack of deep generative networks to learn a mapping from 2D pose to a silhouette with semantic part labels, and per-pixel appearance of the actor. 
We model the person's shape as a dense foreground silhouette mask with per-pixel labels encoding assignment to limbs and garments. 
We further encode the structure of wrinkles and texture patterns of garments as the orientation and strength of local image gradients. 
We extract this structure from images using a bank of oriented filter kernels~\cite{paris2008, tan2020michigan}. 
Encoding the actor's appearance with these explicitly decoupled intermediate representations of silhouette and structure is key to enhance the temporal and spatial quality of synthesized videos comprising human actors in loose clothing. 

Our method improves over current motion transfer approaches in terms of visual fidelity using a single RGB camera. 
Furthermore, our representation provides an additional level of control over the final image generation. For example, for the same overall dynamic geometric outline (\textit{i.e.,} the same garment geometry), color and appearance, including fold and wrinkle style, can be manipulated in a purely image-based way. 
Overall, our contributions can be summarized as follows: 
(1) A new motion transfer framework with an emphasis on visually-plausible fine-scale deformations and dynamics in the actor's clothing. 
(2) For this, we propose to decompose the pose-to-image translation task into better conditioned cascaded processes, where the final appearance is conditioned on the predicted shape outline and internal structure of the clothing.
(3) We show that our intermediate representations do not only help to provide more temporally coherent conditioning resulting in more appealing image synthesis, but also allow controlling individual aspects of the final rendering (\textit{e.g.,} enhance wrinkles and transfer the clothing style). 

\section{Related Work}\label{sec:rw} 

Apart from general-purpose image-translation techniques which can be used for novel human view synthesis~\cite{isola2017image, ZhuPark2017, bansal2018recycle, wang2018pix2pixHD, Wang2018}, most specialized techniques can be classified into still image-based~\cite{Sarkar2020, Siarohin2018, Ma2017, Ma18, Grigorev:2019CVPR} or video-based~\cite{Chan2019, Aberman2018, Esser2018,Liu2019, liu2020neural, lwb2019,Guan2019}. 
Pix2pix~\cite{isola2017image} is one of the most recognized general-purpose image translation techniques in the literature. Other image-to-image translation variants include pix2pixHD~\cite{wang2018pix2pixHD} for processing high-resolution images, Zhu~\etal~\cite{ZhuPark2017} for learning from unpaired data, and Wang~\etal~\cite{Wang2018} for video processing. The work of Bansal~\etal~\cite{bansal2018recycle} learns a mapping from unpaired video data with a focus on processing temporal information. 
While general-purpose image-to-image translation techniques  \cite{isola2017image, ZhuPark2017, bansal2018recycle, wang2018pix2pixHD,Wang2018} can in principle be applied in our  scenario, their visual accuracy usually suffers from multiple  types of artifacts such as spurious details (\textit{e.g.,}  jittering texture details), unnatural deformations, and temporal inconsistency, among others. 

Neural human motion transfer methods are specifically designed to generate novel views of a person observed in a single RGB image.
Still-image-based approaches~\cite{Sarkar2020, Siarohin2018, Ma2017, Ma18, Grigorev:2019CVPR, Esser2018} 
focus is different from the problem we solve. 
Their focus is on re-targeting a person in just a still image and do not produce temporally consistent and visually pleasing video results. 
Video-based human motion transfer techniques~\cite{Chan2019, Aberman2018,Liu2019,liu2020neural, lwb2019,Guan2019}, however, focus on producing video results by handling temporal information.
Aberman~\etal~\cite{Aberman2018} 
combine a skeleton representation with an image-translation network for human performance cloning. 
Additionally, they introduce an actor-independent training branch for unpaired data to enlarge the set of representable poses. 
Recently, Chan~\etal~\cite{Chan2019} introduced a translation method for human bodies using an intermediate, subject independent pose representation.
Inspired by pix2pix~\cite{isola2017image}, they learn a pose-to-appearance mapping from video clips of target identities for human performance transfer. 
To improve temporal consistency and overall image quality, they include a temporal discriminator and dedicated face prediction network.
While their method produces visually appealing results, it frequently generates unnatural and inconsistent deformations and texture details for loose clothing, as it is underconstrained from 2D pose keypoints. 
Some video-based techniques aim to improve temporal and spatial fidelity by incorporating 3D information through human body meshes~\cite{Liu2019,liu2020neural,lwb2019,Guan2019}. 
L.~Liu~\etal~\cite{Liu2019, liu2020neural} presented high-quality human reenactments.
As a proxy, they use an explicitly rigged, textured and skinned 3D model of the target actor. 
The underlying deformation model, however, does not allow to capture small wrinkles and local deformations.
W.~Liu~\etal~\cite{lwb2019} use a human mesh recovery technique to disentangle poses from shapes as well as a correspondence map for human appearance transfer and novel view synthesis. 
They propagate input information both in the image and feature space and advocate a warping-based module for the enhanced preservation of the source information. 
Along the same lines is the work of Guan~\etal~\cite{Guan2019} for human action transfer. 
It shows generalization to different target persons without retraining, thanks to a texture extraction method and a parametric human body model.
However, the resolution of the supported renderings does not capture fine details and wrinkles. 
Textured Neural Avatars~\cite{Shysheya2019} can be trained for rendering a target actor in tight clothes in arbitrary body and camera poses. 
Proceeding from monocular videos with extracted 3D poses and regions of interest (human bodies), they train a convolutional neural network to predict a dense UV-map under a given target pose.
%
%
%
\begin{figure*}[ht]
    \centering
    \includegraphics[width=\textwidth,keepaspectratio]{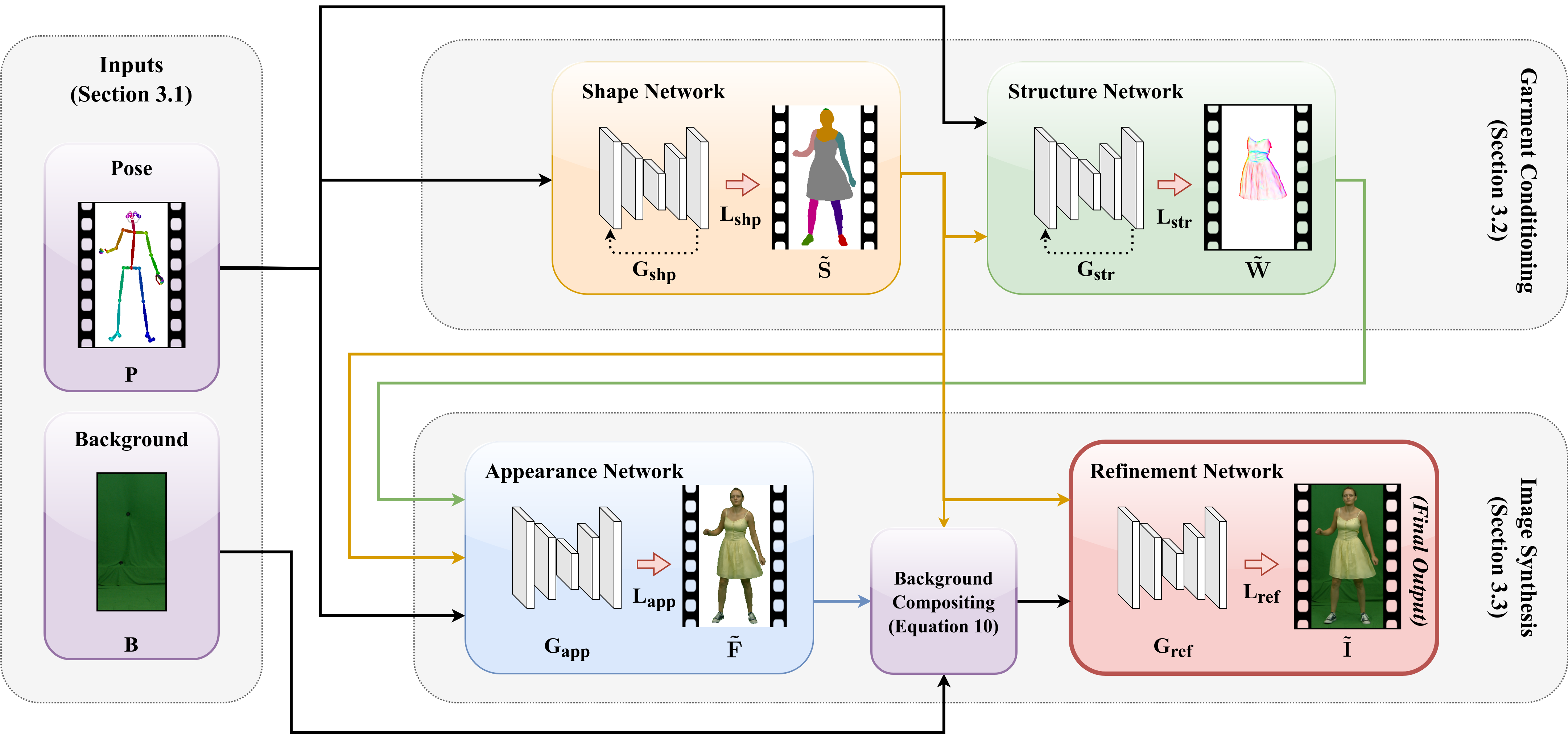}
    \caption{
        \textbf{Framework Overview}: Our framework synthesizes views of humans from a static background image and a pose sequence extracted from the target (training) or source (testing) image sequence.
        Therefore, we apply two consecutive networks to explicitly pre-estimate human body parsing (\textit{yellow}) and the internal gradient structure of clothing (\textit{green}).
        The resulting outputs are used to condition final image generation, which is separated into the  foreground (\textit{blue}) and global (\textit{red}) components to handle loose clothing in a temporally coherent way.
        Dashed lines indicate recurrent networks that feedback the last output to generate the next prediction.
    } 
    \label{fig:framework} 
\end{figure*} 
%
%
%
Explicit decoupling of the geometry and texture results in higher visual quality and temporal consistency; yet, the overall representation is still comparably coarse and lacks fine details as well as local pose-dependent deformations and shading. 
Thus, no body renderings with loose clothes are demonstrated. 
While mesh-based techniques impose stronger human shape priors  for higher quality results, they often assume water-tight  clothing and a pre-captured template of the examined  subject~\cite{Guan2019,Liu2019,liu2020neural,Shysheya2019}. 
This can lead to difficulties in capturing loosely-swinging garments or fine pose-dependent deformations. 
Furthermore, getting access to pre-captured templates can lead to  additional assumptions about the scene (\textit{e.g.,} that the target actor has to stay still for rigid structure-from-motion and brightness consistency for feature point  matching)~\cite{Liu2019,liu2020neural}. 

\section{Method}\label{sec:method}
Our framework $\boldsymbol{G}_\theta$ generates photorealistic videos of a target actor mimicking the motion of another source identity.
Given only a single monocular RGB image sequence $I = (i_n)_{n=1}^{N} \in \mathbb{R}^{N \times 3 \times h \times w}$ containing $N$ frames of width $w$ and height $h$ showing a target actor performing a rich set of motions, we extract an identity-independent 2D skeletal pose $\boldsymbol{P}(i_{n})$ for each frame and fit the network parameters $\theta$ to approximate the inverse mapping $\boldsymbol{P}^{-1}$ back to the original frames. 
Instead of performing the translation task within a single network, our framework can be thought of as a function composition of four generative neural networks ($\boldsymbol{G_{ref}}$, $\boldsymbol{G_{app}}$, $\boldsymbol{G_{str}}$, $\boldsymbol{G_{shp}}$) executed in a cascade fashion to progressively generate higher-level representations, as illustrated in Fig.~\ref{fig:framework}.
Next, we describe all individual framework components (Secs.~\ref{ssec:inputs}-\ref{ssec:image_synthesis}), before providing details on the network architectures and implementation (Sec.~\ref{ssec:implementation}). 

%
\subsection{Inputs}\label{ssec:inputs} 
As inputs, our method takes the target image sequence $I$ and a static image of the scene background $b \in \mathbb{R}^{3 \times h \times w}$ that is fused with the synthesized actor later on. 
From the given image sequence, we extract actor-independent pose representations 
$\boldsymbol{P}: \mathbb{R}^{3 \times h \times w} \rightarrow H \times D$, 
where $H = \{0, 1\}^{C_1 \times h \times w}$ symbolizes the set of rasterized binary pose skeletons with $C_1$ input channels, and $D = \mathbb{R}^{C_2 \times h \times w}$ is  the set of temporal derivatives for a single pose  consisting of $C_2$ channels. 
Similar to recent performance cloning methods \cite{Chan2019,Shysheya2019}, we apply an off-the-shelf pose estimator \cite{cao2021openpose, simon2017hand, wei2016cpm} 
that predicts 2D keypoints $k_n \in \mathbb{R}^{127 \times 2}$ for the body (including hands and face), to generate the skeleton $h_{n} \in H$ by connecting adjacent keypoints via binary lines.
We further distribute $C_1 = 9$ limbs (face, head, torso, arms, legs, hands) over multiple channels, which helps the networks to distinguish between overlapping or symmetrical body parts. 
However, estimating soft body deformations from a single pose remains highly ambiguous as the states depend on the temporal order of poses. 
Thus, we calculate a temporal pose context $d_{n} \in D$ as the first and second temporal derivatives of $k_{n}$ with respect to the image index $n$ in x and y direction,  respectively ($C_2 = 2 \cdot 2 \cdot 9$), similar to velocities and accelerations in classical dynamics simulation: 
\begin{equation} \label{eq:dynamics}
    k'_{n} = \left(\frac{\partial k_{n}}{\partial n}, \frac{\partial^2 k_{n}}{\partial n^2}\right).
\end{equation} 
We create rasterizations $d_{n}$ from $k'_{n}$ in the same way as for the pose skeletons, and linearly interpolate their values along the bones, similar to the depth representation of  Shysheya~\etal~\cite{Shysheya2019}. 
During the reenactment, we apply the same procedure to the source image sequence, but additionally perform pose normalization as described by Chan~\etal~\cite{Chan2019} for inter-target appearance transfer before drawing the skeletons. 
Finally, our pose conditioning $\boldsymbol{P}(i_{n})=(h_{n}, d_{n})$ is provided to our framework as the concatenation of the framewise pose skeleton and  temporal context. 
We find that the described procedure results in a compact, yet expressive pose representation, while slightly outperforming a simple sliding window approach in our experiments. 

\subsection{Garment Conditioning}
In our framework, we use two dedicated networks that pre-estimate the actor's shape and the internal gradient structure of clothing in 2D space to improve the consistent modeling of deforming garments and the fall of the folds. 
%
\subsubsection{Shape Estimator}
As a first step, our shape estimator $\boldsymbol{G_{shp}}$ transforms the provided pose representations $\boldsymbol{P}$ into the actor's current silhouette
\begin{equation} \label{eq:shape_estimator}
        \tilde{s}_{n}^{*} = \boldsymbol{G_{shp}}(\boldsymbol{P}(i_{n}), \tilde{s}_{n-1}^{*}), \\
\end{equation}
where $\tilde{s}_{i}^{*} \in \mathbb{R}^{j \times h \times w}$ is represented as semantic body-part segmentation with $j$ labels.
We extract a final segmentation mask $\tilde{s}_{n} \in ([\mathbb{N}]_{1}^{j})^{h \times w}$ from our network by extracting the indices of maximum elements per pixel over the channel dimension. 
As knowledge about the preceding shape is necessary to achieve temporal coherence of deforming garments, we design our networks in a recurrent fashion where each execution is conditioned on the previous output. 
Thus, $\boldsymbol{G_{shp}}$ is tasked to estimate a delta in the observed shape  based on the current pose and the contextual temporal derivatives. 
The resulting segmentation mask is a significant prior for the final appearance network to produce visually-plausible renderings, and it is further used to segment the actor and parts of their clothing in later stages. 
For training, we use the human parsing method of Li~\etal~\cite{li2019self} trained on the ATR dataset \cite{liang2015deep} to extract pseudo-ground-truth segmentation maps $S = (s_n)_{n=1}^{n}$ from $I$.
This dataset provides $j=18$ labels for all body limbs and common clothing, enabling our method to handle various clothing styles including shirts, pants, skirts, dresses and scarfs.
We formulate our shape training loss $\boldsymbol{L_{shp}}$ as the cross-entropy between the network prediction and the training label: 
\begin{equation} \label{eq:loss_shape_network}
    \boldsymbol{L_{shp}} = \mathbb{E}_{n \sim N} \log\left(\sum_{j}\exp(\tilde{s}_{n}^{*}(j))\right) - \tilde{s}_{n}^{*}(s),
\end{equation}
where $\tilde{s}_{n}^{*}(j)$ is the output channel for a given label $j$.
\subsubsection{Structure Estimator}
While our shape representation is sufficient prior to infer the appearance of solid body parts like arms and faces, 
it does not provide information about wrinkles and folds within the clothing, that are needed to generate temporally consistent shading and texture patterns.
Thus, we apply a second recurrent network $\boldsymbol{G_{str}}$ that estimates the internal gradient structure $\tilde{w}_{n} \in \mathbb{R}^{2 \times h \times w}$ for clothing regions as indicated by the segmentation map $\tilde{s}_{n}$:
\begin{equation} \label{eq:structure_network}
    \tilde{w}_{n} = \boldsymbol{G_{str}}(\boldsymbol{P}(i_{n}), \tilde{s}_{n}, \tilde{w}_{n-1}).
\end{equation}
We model the clothing structure as the pixelwise gradient direction and strength extracted from the responses of $32$ oriented Gabor filters, similar to recent work on neural hair synthesis \cite{tan2020michigan}. 
However, as a floating garment comprises sparser gradients than human hair, we do not only use the maximum filter responses to smooth the angle field, but append a normalized version to the final gradient directions to model the probability of a wrinkle or texture change at a specific location, as visualized in Fig.~\ref{fig:teaser} (color and saturation encode the gradient direction and the strength, respectively). 
We train the structure estimator using the $L_1$ distance between estimated $\tilde{w}_{n}$ and ground-truth $w_{n}$ structure using a mask for garment labels extracted from $\tilde{s}_n$: 
\begin{equation} \label{eq:structure_loss}
\small
        \boldsymbol{L_{str}} = \mathbb{E}_{n \sim N} \: \chi_C(\tilde{s}_n) \, |\tilde{w}_{n} - w_{n}|,
\end{equation}
where $C$ denotes the set of segmentation labels corresponding to the actor's clothing, and $\chi_C$ is the indicator function: 
\begin{equation}
    \chi_C(x) =  \begin{cases} 1, \, &\mbox{if} \, x \in C, \\ 0, \, &\mbox{else}. \end{cases} 
\end{equation} 

%
\subsection{Image Synthesis}\label{ssec:image_synthesis} 
The second stage of our framework synthesizes the final output image based on the provided pose and garment conditionings. Again, we use two dedicated networks to independently generated the actor in the foreground and fuse it with the provided background image.
%
\subsubsection{Appearance Network}
Our first rendering network $\boldsymbol{G_{app}}$  takes provided pose as well as the pre-estimated  shape $\tilde{S}$ and internal clothing structure  $\tilde{W}$ to synthesize the actor's appearance  $\tilde{f}_{n} \in \mathbb{R}^{3 \times h \times w}$: 
\begin{equation} \label{eq:appearance_network}
    \tilde{f}_{n} = \boldsymbol{G_{app}}(\boldsymbol{P}(i_n), \tilde{s}_{n}, \tilde{w}_{n}).
\end{equation}
We train the appearance module using a combination of $L_1$ distance and perceptual reconstruction loss \cite{johnson2016perceptual}. 
Again, an indicator function $\chi_B$ is used to mask the foreground pixels that are not assigned to the  background label $\beta$: 
\begin{equation} \label{eq:fg_mask}
    \chi_B(x) =  \begin{cases} 0, \, &\mbox{if} \, x = \beta, \\ 1, \, &\mbox{else}. \end{cases} 
\end{equation}
Thus, our appearance loss $\boldsymbol{L_{app}}$ reads
\begin{equation} \label{eq:loss_appearance_network}
    \boldsymbol{L_{app}} = \mathbb{E}_{n \sim N} \: \chi_B(\tilde{s}_n) \, \left( \lambda_{r}|\tilde{f}_{n} - i_{n}| + \lambda_{p}|\phi(\tilde{f}_{n}) - \phi(i_{n})| \right),
\end{equation}
where $\phi(\cdot)$ denotes feature maps extracted from different layers of a pre-trained VGG19 network \cite{simonyan2014very}, and $\lambda_{r}, \lambda_{p}$ are free hyperparameters for weighting.
%
\subsubsection{Refinement Network}
Finally, we apply a shallow refinement network to fuse the foreground prediction with the provided scene background.
For this, we first paste the generated foreground $\tilde{f}_n$ onto the static background image $b$ using the masking function \eqref{eq:fg_mask}:
\begin{equation} \label{eq:background_compositing}
    \tilde{i}^{*}_{n} = (\chi_B(\tilde{s}_n) \cdot \tilde{f}_{n}) + ((1-\chi_B(\tilde{s}_n)) \cdot b).
\end{equation}
Then, given foreground-background composition $\tilde{i}^{*}_{n}$ and actor segmentation $\tilde{s}_{n}$, our refinement network $\boldsymbol{G_{ref}}$ performs simple transition smoothing and shadow generation to produce the final output image $\tilde{i}_n$:
\begin{equation} \label{eq:refinement_network}
    \tilde{i}_{n} = \boldsymbol{G_{ref}}(\tilde{i}^{*}_{n}, \tilde{s}_{n}).
\end{equation}
We use a combination of structural and perceptual losses similar to our $\boldsymbol{L_{app}}$ \eqref{eq:loss_appearance_network}, but over the entire image plane: 
\begin{equation} \label{eq:loss_refinement_network}
    \boldsymbol{L_{ref}} = \mathbb{E}_{n \sim N} \: \lambda_{r}|\tilde{i}_{n} - i_{n}| + \lambda_{p}|\phi(\tilde{i}_{n}) - \phi(i_{n})|.
\end{equation}
In contrast to recent pose-to-video translation methods \cite{Aberman2018, Chan2019}, we do not apply an adversarial loss for our generator network as we do not want to hallucinate high frequency details in the clothing based on statistics from the data sequence $I$, but rather encourage the network to stick to the predicted structure layout.
%
\subsection{Implementation Details}\label{ssec:implementation} 
We implement our framework in Python using PyTorch 1.5 \cite{NEURIPS2019_9015}.
We intend to release the full implementation,  including data sequences to reproduce the presented  results. 
Our four translation networks employ the local Pix2PixHD \cite{wang2018pix2pixHD} generator architecture; we halve the number of residual blocks in our refinement network $\boldsymbol{G_{ref}}$ due to the similarity of the domains.
We optimize our final loss
\begin{equation} \label{eq:loss_total}
    \boldsymbol{L_{total}} = \lambda_{1}\boldsymbol{L_{seg}} + \lambda_{2}\boldsymbol{L_{str}} + \lambda_{3}\boldsymbol{L_{app}} + \lambda_{4}\boldsymbol{L_{ref}}
\end{equation}
with hyperparameters $(\lambda_{1} = 0.5,  \lambda_{2,3,4} = 1.0, \lambda_{r} = 0.1, \lambda_{p} = 0.9)$ in a sequential way for $30$ epochs using Adam optimization \cite{kingma2014adam} (momentum $\beta_1 = 0.999$ and $\beta_2 = 0.5$).
On a single Quadro RTX 8000 GPU, training converges after approximately four days for $N \approx 23K$ frames, while processing a single testing-frame takes ${\sim}280$ milliseconds at a resolution of $512 \times 512$ pixels. 

\section{Experiments}\label{sec:eval}
%
%
%
\begin{figure*}[ht!]
    \centering
    \includegraphics[width=\textwidth, keepaspectratio]{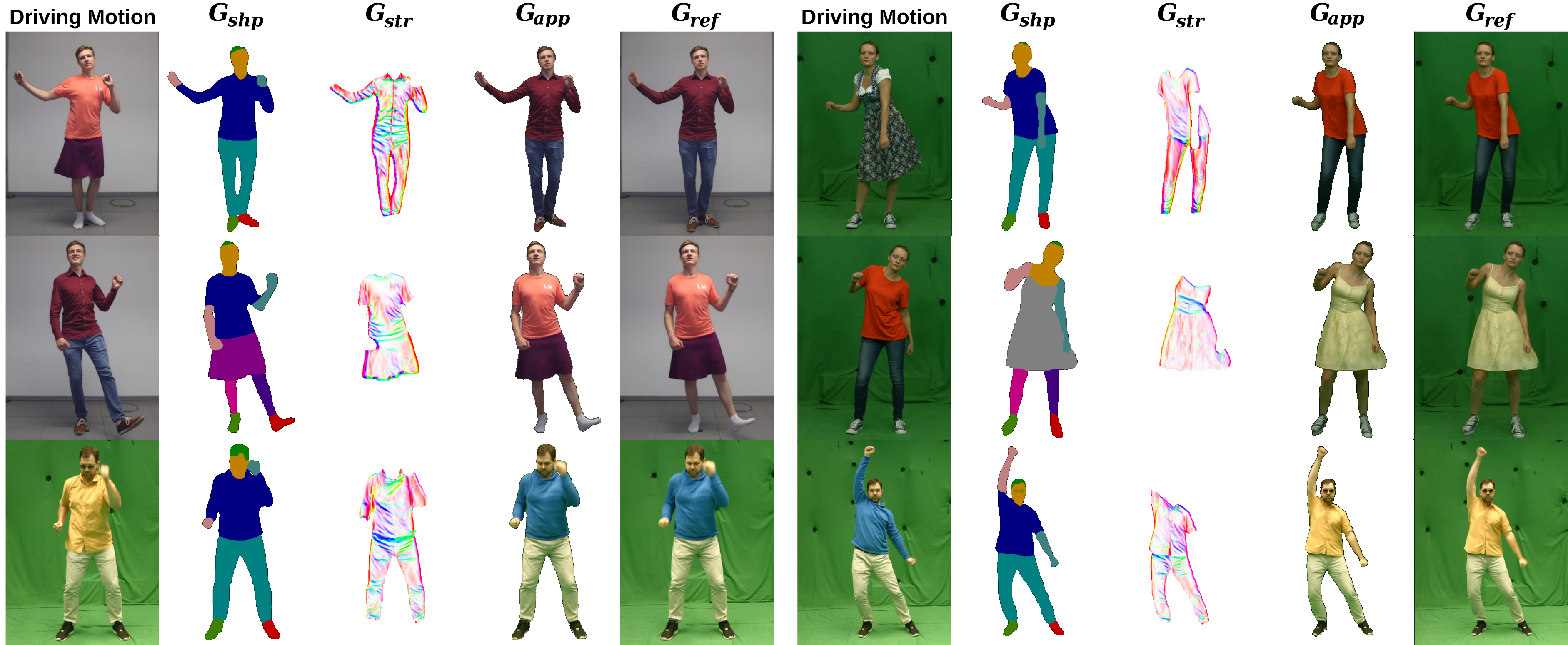}
    \caption{
    \textbf{Exemplary results of our motion transfer framework}. The first column of each block illustrates the driving sequence, while the last one shows the final result. They anchor the outputs for the different intermediate networks (shape, structure, and appearance). For clothing gradient structure, we encode the direction and confidence as color and saturation,  respectively, similar to optical flow visualization.} 
    \label{fig:ourResults}
\end{figure*}
%
%
%
We compare our framework to several state-of-the-art approaches
with subjects wearing different types of loose clothing with rich wrinkling and textures (Sec.~\ref{ssec:results_comparisons}), conduct a user study (Sec.~\ref{ssec:perceptual_experiment}) and perform an ablation study for individual framework stages  (Sec.~\ref{ssec:ablation_study}). 
Since our results improve upon the temporal consistency and fine texture details, they are best viewed in our supplementary video. 

\subsection{Datasets, Methods and Metrics}
Existing datasets for the evaluation of recent appearance transfer methods usually comprise actors in monochromatic body-tight clothing, as loose garments are a frequent cause of visual artifacts and temporal irregularities. 
For our experiments, we capture new sequences of three subjects, each in up to three clothing styles with a length of seven to ten minutes at a resolution of $512 \times 512$ pixels, and select the last $~5\%$ of the frames for testing. 
Our dataset features various clothing styles including jeans, loose t-shirts, hoodies, and two dresses with rich wrinkle and texture patterns. 

We compare our framework to current video-based appearance transfer approaches, \textit{i.e.}, a general-purpose pix2pixHD network~\cite{wang2018pix2pixHD} that synthesizes an image from a rendering of the extracted pose, the Everybody-Dance-Now (EDN) method by Chan~\etal~\cite{Chan2019}, which adds a temporal discriminator and specialized face GAN, and  Recycle-GAN~\cite{bansal2018recycle} that performs direct unsupervised retargeting between two videos. 
Additionally, we test our approach on the dataset of Liu~\etal~\cite{liu2020neural} comprising various actors in tight clothes. 
All results were generated using either the official implementations or data directly provided by the authors.
Moreover, we have explored pose editing using NHRR~\cite{Sarkar2020} and Liquid Warping GAN~\cite{lwb2019} but the results showed poor visual quality on our sequences. 
Kindly note that these methods predict pixel correspondences to the SMPL mesh~\cite{SMPL:2015} using DensePose~\cite{guler2018densepose}, and hence struggle with handling loose clothes.
We provide exemplary results showing limitations of mesh-based approaches in our supplementary material.
Furthermore, the code of Textured Neural Avatars~\cite{Shysheya2019} is, unfortunately, not publicly available, and we were not able to obtain results on our sequences directly from the authors. 
For quantitative analysis, we utilize the widely-used SSIM \cite{wang2004image}, LPIPS \cite{zhang2018perceptual} and FID \cite{heusel2017gans} metrics that assess the pixel-space structural similarity, perceptual distance based on neural network features, and the Fr{\'e}chet distance between two data collections, respectively. 

\subsection{Results and  Comparisons}\label{ssec:results_comparisons} 
\begin{figure}[bt!]
    \centering
    \includegraphics[width=\columnwidth, keepaspectratio]{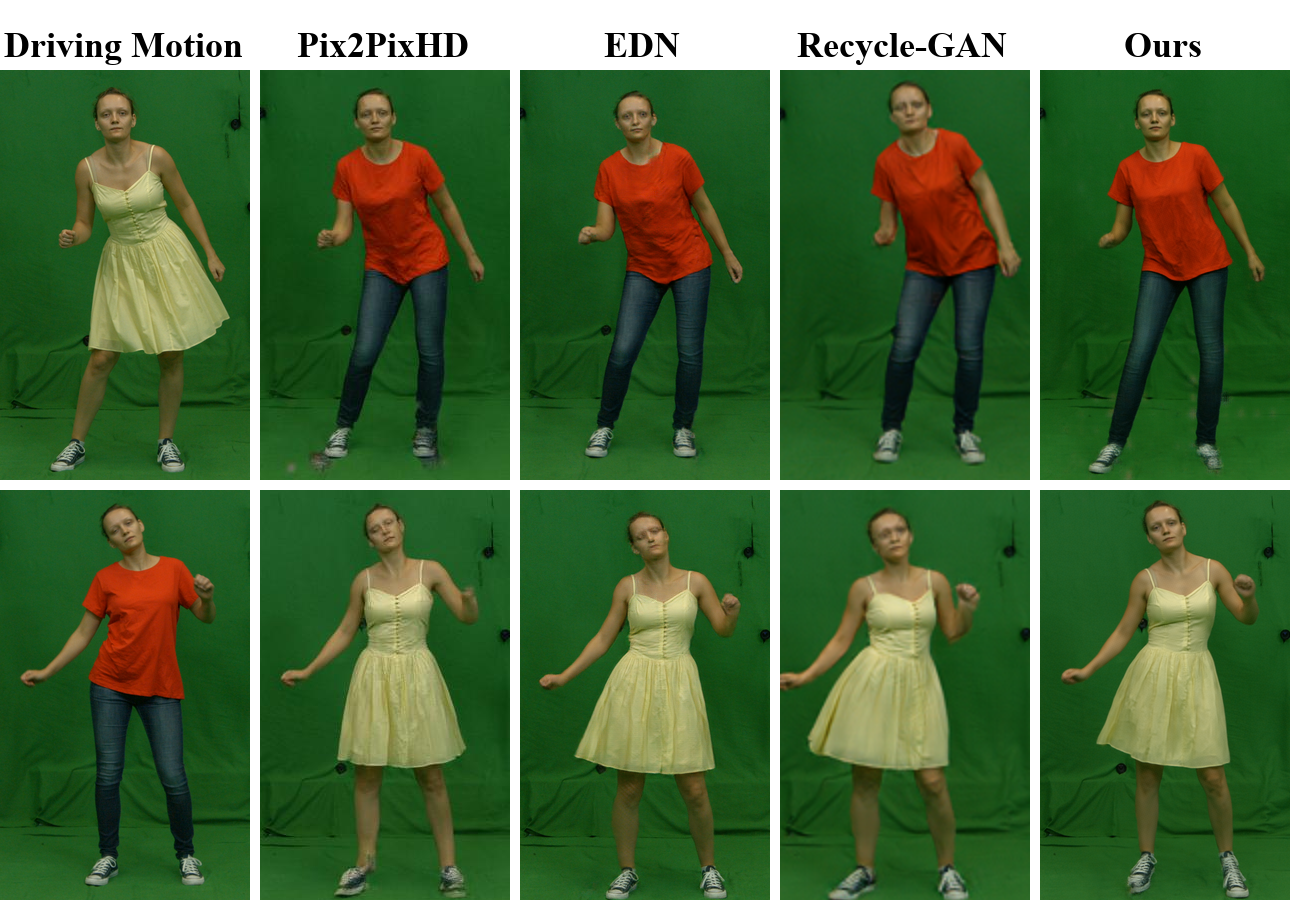}
    \caption{
    \textbf{Comparison to state-of-the-art} motion transfer methods. We compare our method against pix2pixHD~\cite{wang2018pix2pixHD}, EDN~\cite{Chan2019} and Recycle-GAN~\cite{bansal2018recycle} on two sequences of an actress in a loose t-shirt and a dress. Please note that results of Recycle-GAN are only available at half of the original resolution, as training on full resolution produced highly implausible results.} 
    \label{fig:sota}
\end{figure}

We show exemplary results of every network component in Fig.~\ref{fig:ourResults}. 
As can be seen, our framework is capable of estimating pronounced body part segmentations, the corresponding  wrinkle patterns and resulting  target actor appearance for all types of complex clothing in our dataset, independent of the source actor's posture or appearance.
%
We further compare our method to recent state-of-the-art approaches in Fig.~\ref{fig:sota}.
Therefore, we use a bidirectional reenactment scenario with two different types of clothing, \textit{i.e.,} a loose monochromatic t-shirt with jeans, and a dress featuring a delicate arrangement of folds. 
While all of the tested methods are able to produce strong wrinkle patterns, our framework manages to achieve the most plausible temporal and spatial results  due to the explicit handling of deforming garments. 
Please note that the temporal consistency can only be evaluated by watching our supplemental video.
Also, we provide results and comparisons on the video sequences of Liu~\etal~\cite{liu2020neural} in our supplemental material.

%
%
%
%
%
We further assess the visual quality by numerically comparing our method to related pose-to-video translation approaches in Table~\ref{tab:quantitative_results}. 
To calculate the metrics, we employ the target actor sequences shown in the first column of Fig.~\ref{fig:sota} using the corresponding self-reenactment scenarios, where ground-truth images are available from the testing sets. 
\begin{table}[th!]
    \centering
    \resizebox{1.0\linewidth}{!}{%
    \begin{tabular}{ | p{0.25\linewidth} p{0.05\linewidth} | p{0.3\linewidth} | p{0.2\linewidth}  | p{0.2\linewidth} |}  
    \hline
    Metric &  & pix2pixHD \cite{wang2018pix2pixHD} & EdN \cite{Chan2019} & ours \\ \hline
    SSIM \cite{wang2004image} & $\uparrow$ & 0.9288 & 0.9043 & \textbf{0.9358}  \\ 
    LPIPS \cite{zhang2018perceptual} & $\downarrow$ & 0.0401 & 0.0303 & \textbf{0.0289}  \\
    FID \cite{heusel2017gans} & $\downarrow$ & 21.7724 & \textbf{13.5421} & 17.9573 \\ 
    \hline
    \end{tabular}}
    \caption{
    \textbf{Quantitative comparison} to related methods. We assess the quantitative quality according to SSIM, LPIPS and FID metrics in a self-reenactment scenario.
    } 
    \label{tab:quantitative_results}
\end{table}
Recycle-GAN~\cite{bansal2018recycle} focuses on learning from unpaired data and is upper-bounded by paired translation techniques when trained on paired data. Hence, we exclude it from our quantitative analysis. 
Our method outperforms the current state of the art according to the SSIM and LPIPS metrics, which perform a direct structural and perceptual comparison on a per-frame basis, indicating that the physical behaviour of clothing synthesized by our method better reflects ground truth.
On the other hand, FID states that the pixel distribution generated by EDN is closer to the training set, presumably due to strong adversarial losses.
%
%
\subsection{Perceptual  Experiment}\label{ssec:perceptual_experiment} 
We find that quantitative metrics do not reveal all aspects of the visual video quality, and complement our evaluation by two user studies in which videos are graded according to human perception. 
We compare our method against recent image-based techniques using the video sequences shown in Fig.~\ref{fig:sota}, including wide clothes with significant deformations. We also conduct an additional experiment on the dataset of Liu~\etal~\cite{liu2020neural} to evaluate on videos of actors in tight clothes.  
In both studies, we show the participants video reenactments of ${\sim}20$ seconds in length with results of two methods at a time and ask them which video looks the most realistic. 
%
%
\begin{table}[b!]
    \centering
    \setlength\tabcolsep{2 pt}
    {\footnotesize
    \begin{tabular}{|c|l|l|c|c|}
         \hline
         Exp.&Method &Input & \#Votes & Ranking\\
         \hline
         \multirow{4}{*}{11}&\textbf{Ours} &Video (Pose) & 257&\textbf{1}\\
         &Recycle-GAN~\cite{bansal2018recycle} &Video (RGB) & 194 &\textbf{2}\\
         &EDN~\cite{Chan2019} & Video (Pose)& 143&\textbf{3}\\
         &pix2pixHD~\cite{wang2018pix2pixHD} &Video (Pose)& 54&\textbf{4}\\
          \hline
          \hline
         \hline
        \multirow{3}{*}{2}&\textbf{Liu}~\cite{liu2020neural} & Textured mesh + Video (Pose) & 103& \textbf{1}\\
         &Ours & Video (Pose) & 76 & \textbf{2}\\
         &EDN~\cite{Chan2019} &Video (Pose) & 13 & \textbf{3}\\
         \hline
        \end{tabular}
    }
    \caption{\textbf{Perceptual ranking} of the compared methods for our user studies. Experiment 1 was conducted online with 
    $54$ participants with various backgrounds, while experiment 2 was in-situ with $16$ CG/CV experts. The rankings are statistically significant.}
    \label{tab:votesExp}
\end{table}
To measure both the performance of each method and the agreement between  participants, we follow the linked-paired comparison  design~\cite{david1963method}. 
We rank each method according to the number of times they are preferred over the rest and perform a significance test of the votes differences~\cite{Setyawan:2004,Castillo:2011}. 
The total number of votes a method received in each experiment is displayed in Table~\ref{tab:votesExp}. 
As can be seen, our method significantly outperforms recent  image-based pose-to-video-translation methods in terms of video realism for the clothing and is even preferred over Recycle-GAN, which uses a richer input domain for motion transfer. 
Our second experiment on the dataset of Liu~\etal~\cite{liu2020neural} further confirms that our monocular video-based approach also achieves considerably more realistic results than the related EDN method when exchanging motion between actors in tight clothing.
However, in that scenario, image-based approaches cannot quite reach the quality of priors provided by expensively recorded template meshes, which excel for tight and static shapes (Liu~\etal~\cite{liu2020neural}). 
We provide more details on the experimental setup and evaluation in our supplementary material. 
%
%
\begin{figure}[tb!]
    \centering
    \includegraphics[width=\columnwidth,keepaspectratio]{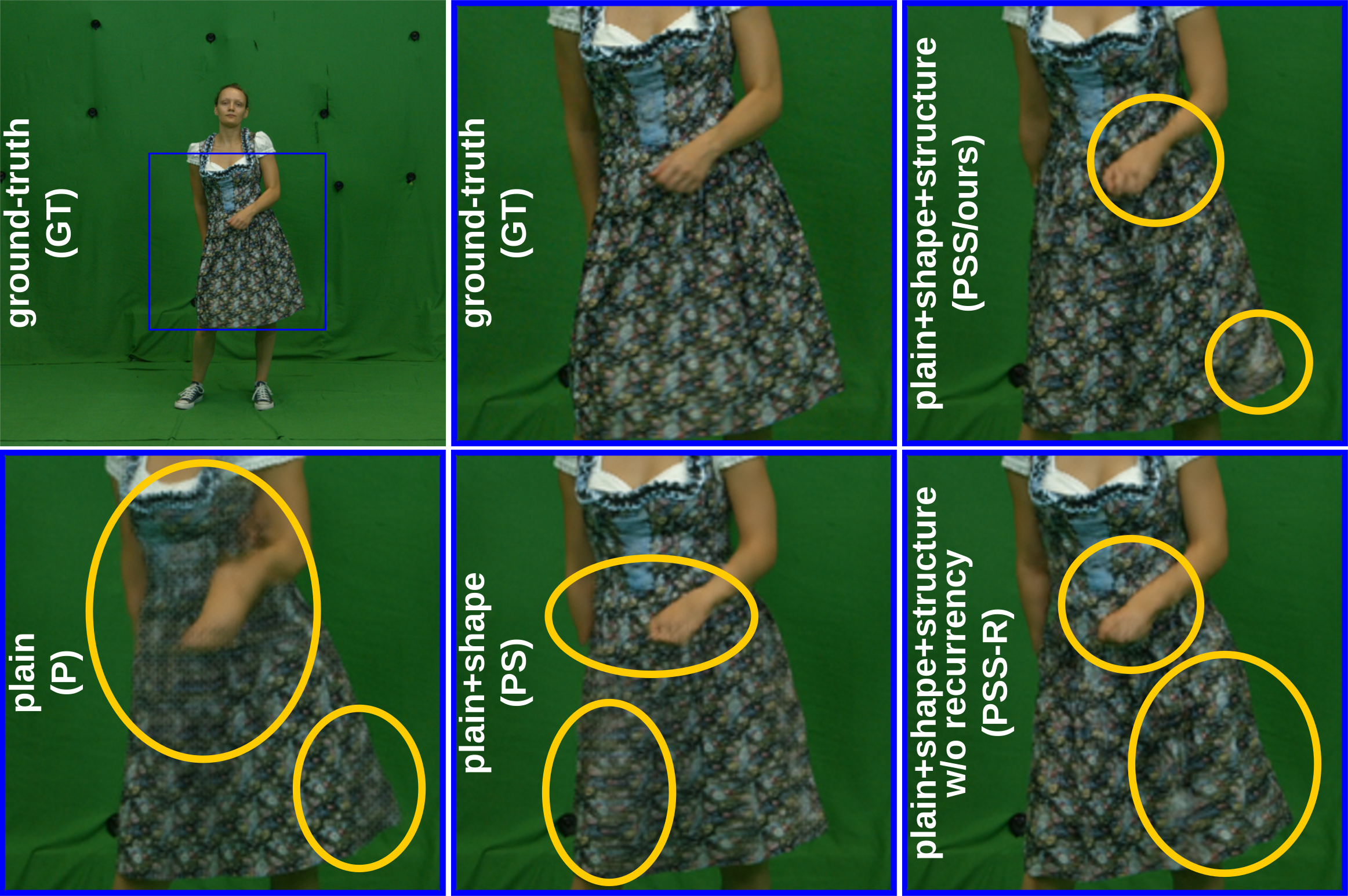}
    \caption{\textbf{Ablation study} evaluating the influence of our framework stages in a self-reenactment scenario.
    For an exemplary ground-truth frame, we show a plain appearance network conditioned on pose only (P), pose and shape (PS), and our full framework without (PSS-R) and with (PSS/ours) temporal recurrence. 
    } 
    \label{fig:ablation} 
\end{figure} 
%
\subsection{Ablation Study}\label{ssec:ablation_study} 
We next conduct an ablation study to assess the influence of different framework components. 
Therefore, we retrain our method on one of our most challenging sequences featuring a swinging dress with complicated texture patterns, while incrementally dropping the garment conditioning networks. 
In our study, we consider four framework variations, \textit{i.e.,} the plain appearance network without additional inputs (P), the appearance conditioned on our shape estimates (PS) and our full framework including the internal structure (PSS/ours); additionally, we examine the effect of removing the recurrent back-feeding from all network modules (PSS-R). 
As shown in Fig.~\ref{fig:ablation}, we find that every  component contributes to improving the final image quality. 

While our plain pix2pix generator fails to produce sharp outlines of body parts and garments (P), 
%
%
%
\begin{figure*}[th!]
    \centering
    \includegraphics[width=\textwidth,keepaspectratio]{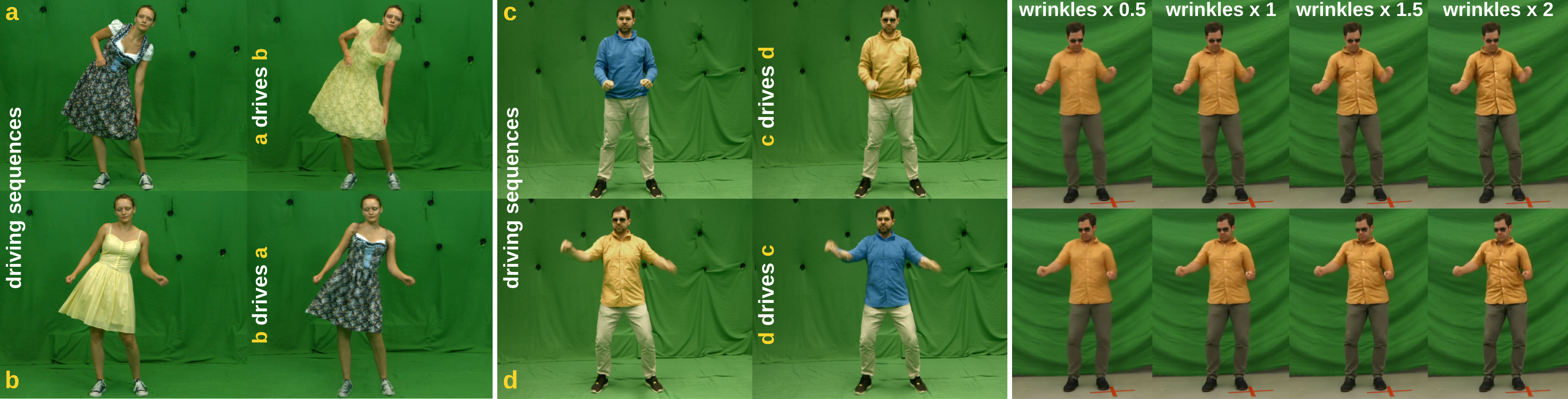}
    \caption{\textbf{Applications:} Our approach allows novel video editing through style transfer and wrinkle intensity manipulation. For style transfer, we use the source shape and/or structure while keeping the target appearance. 
    }
    \label{fig:apps}
\end{figure*}
%
%
\begin{table}[hb!]
    \centering
    \resizebox{1.0\linewidth}{!}{%
    \begin{tabular}{ | p{0.2\linewidth} p{0.02\linewidth} | p{0.18\linewidth} | p{0.18\linewidth}  | p{0.18\linewidth} | p{0.18\linewidth} |}  
    \hline
    Metric &  & P & PS & PSS-R & PSS (ours) \\ \hline
    SSIM \cite{wang2004image} & $\uparrow$ & 0.8959 & 0.9045 & 0.9045 & \textbf{0.9046}  \\ 
    LPIPS \cite{zhang2018perceptual} & $\downarrow$ & 0.044 & 0.036 & 0.036 & \textbf{0.035}  \\
    FID \cite{heusel2017gans} & $\downarrow$ & 15.046 & 12.282 & 12.925 & \textbf{12.176} \\ 
    \hline
    \end{tabular}}
    \caption{\textbf{Quantitative ablation} analysis according to SSIM, LPIPS and FID metrics. We compare different versions of our framework in a self-reenactment scenario. The numbers indicate that every component contributes a structural and perceptual improvement to the synthesized output.} 
    \label{tab:ablation}
\end{table}
providing a pre-estimated shape mask results in distinct transitions and more realistically-floating garments (PS). 
Adding information about the internal structure of the clothing further enhances the generation of high-frequency details within the given shape (PSS). 
This effect is also observable in quantitative evaluation using structural and perceptive metrics, as shown in Table~\ref{tab:ablation}.
Furthermore, we observe that dropping the temporal back-feeding of previous network outputs (PSS-R)---while achieving similar quality on a per-frame basis--- significantly decreases temporal consistency, as can be seen in our supplemental video. 
\section{Application: Material Editing}
On top of high-fidelity motion transfer, our framework offers an additional level of control over the image generation, as it incorporates explicit specifications of the actor's shape and garment structure. 
This enables editing and manipulating the generated videos. 
For actors wearing similar clothing, pseudo-ground-truth segmentation or structure maps extracted from the source sequence can significantly improve the final image quality and temporal consistency. 
Alternatively, after training multiple subjects wearing clothing of the similar type, garment conditioning networks can be interchanged to generate clothing with new combinations of shape, structure and appearance. 
Furthermore, manually-edited or completely handcrafted shape and structure patterns can be used to influence the physical behavior of clothing. 
Fig.~\ref{fig:apps} shows two examples of automatic and manual material and appearance editing. 
We change the style of the actress's dress and the actor's top while keeping the original appearance by applying the shape (left), or both shape and structure, from the source sequence and feed it to the appearance network (middle). 
Also, magnifying the second channel of structure estimate (gradient confidence map) can be used to magnify or minify the fall of the folds during reenactment (right). 
While material property exchange is still currently limited to garments of comparable type or texture, the proposed approach is the first neural motion transfer method enabling controllable clothing composition for videos, which can pave the way for many applications in future (such as virtual try-on).  
\section{Discussion and Limitations}\label{sec:dis} 

We have demonstrated high-fidelity results for multiple challenging human appearances with various deformation patterns of garments, and were able to manipulate several properties of the clothing during the reenactment. 
The ablative study has shown that all stages of our framework are justified and contribute to the final result. 
Although our method is a step forward in the neural rendering of humans, it has limitations and can provoke further research. 
First, the space of depictable poses is currently determined by the composition and length of the training sequences. 
Next, inaccuracies and missing 3D information from the pose keypoint estimator can result in incorrect occlusion maps and temporal behavior of garments. 
At the same time, the propagation of errors in the estimated human body parsing can lead to missing or disconnected body parts and unnatural transitions between the frames. 
These aspects can be improved in future by incorporating stronger constraints on the shape and body structure and enlarging the pose space by training on unpaired data. 

\section{Conclusion}\label{sec:conclusion} 
We introduced a novel multi-stage framework for high-fidelity human motion transfer from monocular video. 
The tests showed that our method can perform qualitatively appealing motion transfer for different clothes, improving over previous image-based techniques in visual quality. 
The performance is also confirmed by a comprehensive perceptual study, indicating superior temporal and visual quality compared to current image-based state-of-the-art motion transfer approaches. 
All in all, we conclude that explicitly handling deformations of clothing in a temporally-coherent way is crucial for high-fidelity neural human motion transfer.

\section*{Acknowledgements}\label{sec:acknowledgements}
The authors gratefully acknowledge funding by the German Science Foundation (DFG MA2555/15-1 ``Immersive Digital Reality''). 
This work was partially funded by the ERC Consolidator Grant 4DRepLy (770784). 
We thank Jalees Nehvi and Navami Kairanda for helping with comparisons. 
%

%
{\small 
\bibliographystyle{ieee_fullname} 
\bibliography{references} 
} 
\twocolumn[{ 
\renewcommand\twocolumn[1][]{#1} 
\centering
\Large{\textbf{High-Fidelity Neural Human Motion Transfer from Monocular Video \\ --- Supplementary Material ---}} %
\vspace{30pt}
}] 
\setcounter{section}{0}
\setcounter{figure}{0}
\appendix
\renewcommand{\thetable}{\Alph{table}}
\renewcommand{\thefigure}{\Roman{figure}}
In this appendix, we provide more details on the methods examined in the main paper along with additional results for the presented framework. 
\section{Garment Conditioning Representation and Visualization} 
\begin{figure}
    \centering
    \includegraphics[width=\columnwidth,keepaspectratio]{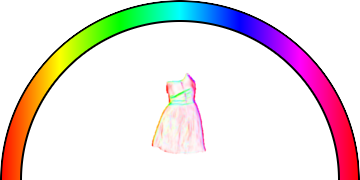}\\
    \caption{Example of our garment structure visualization for one of the dresses shown in Fig.~\ref{fig:teaser} from the main paper. We model the orientation as hue in HSV-color space, while each pixel's saturation corresponds to the local gradient confidence.}
    \label{fig:colorwheel}
\end{figure}

Our current implementation uses the self-correction for human parsing method by Li~\etal~\cite{li2019self}, that was trained on the ATR dataset~\cite{liang2015deep}. 
Thus, our method currently supports a total of eighteen different labels (namely background, hat, hair, sunglasses, upper clothes, skirt, pants, dress, belt, left shoe, right shoe, face, left leg, right leg, left arm, right arm, bag and scarf). 
To generate training data and visualize the results, we use the official code of \cite{li2019self} provided by the authors\footnote{\url{https://github.com/PeikeLi/Self-Correction-Human-Parsing}}. 
Naturally, our framework is compatible with arbitrary human parsing methods, and can easily be modified to support the latest state-of-the-art approaches as well as new datasets including labels for different types of clothing. 
To estimate the internal gradient structure, we adopt the procedure described by Tan~\etal~\cite{tan2020michigan} for conditioning hair structure. 
More precisely, we extract ground-truth annotations from the video sequence using a set of 32 oriented Gabor-filters $K_{\Theta}$ with $\Theta = [0, \pi)$ being the discrete angle values. 
By applying the filter stack to every pixel, we extract a dense orientation $o_n$ and confidence $c_n$ map for image $i_n$ by calculating the angle and amplitude of the maximum filter response as: 
\begin{equation} 
\begin{aligned} 
o_n =& \argmax_{\Theta}\vert \left( K_{\Theta} \otimes i_n  \right) \vert, \\ 
c_n =& \max_{\Theta}\vert \left( K_{\Theta} \otimes i_n  \right) \vert, \\ 
\end{aligned} 
\end{equation} 
where $\otimes$ denotes the convolution operator.
We then follow their procedure of converting $o_n$ to a continuous representation, and Gaussian-filter the orientation map based on the local confidence $c_n$ to reduce noise. 
However, in contrast to hair, which comprises a dense structure field by nature, the local gradient structure of clothing is usually sparse (\textit{e.g.,} tight monochromatic cloth) and highly dependent on the texture and material. 
Thus, unlike the original implementation that discards confidence after filtering, we normalize $c_n$ in the range $[0, 1]$, and append it to the orientation map, which results in our final two-channel clothing structure representation $w_n = (o_n, c_n)$.
Thus, intuitively, the first channel of our clothing structure tells the appearance network in which direction a gradient (\textit{e.g.,} produced by a wrinkle) points, while the second channel models the confidence and strength of texture changes for each individual pixel. 
In our figures and supplemental video, we visualize the structure representation $w_n$ similar to optical flow vectors, where the orientation and length (confidence) are modeled as hue and saturation in HSV-color space,  respectively, as shown in Fig.~\ref{fig:colorwheel}. 
%
\section{Perceptual Experiment}\label{sec:experiment}
In our perceptual experiments, we aim at comparing the reenacting results from an observer’s perspective, which requires multiple stimuli with differences between them often being quite subtle. More importantly, the quality of the results that we aim to measure cannot be represented on a linear scale \cite{Kendall1940}, which advises against ranking the methods. Therefore, we chose the paired comparisons technique, where the participants are shown two reenacted videos at a time, side by side, and are asked to choose the one that better fits the task question. We thus performed a two-alternatives-forced-choice (2AFC) preference task assessing reenactments by two compared methods for a given video. 

\subsection{Stimuli}
\begin{figure}
    \centering
    \includegraphics[width=\columnwidth,keepaspectratio]{images/sota.png}\\
    \vspace{5pt}
    \includegraphics[width=\columnwidth,keepaspectratio]{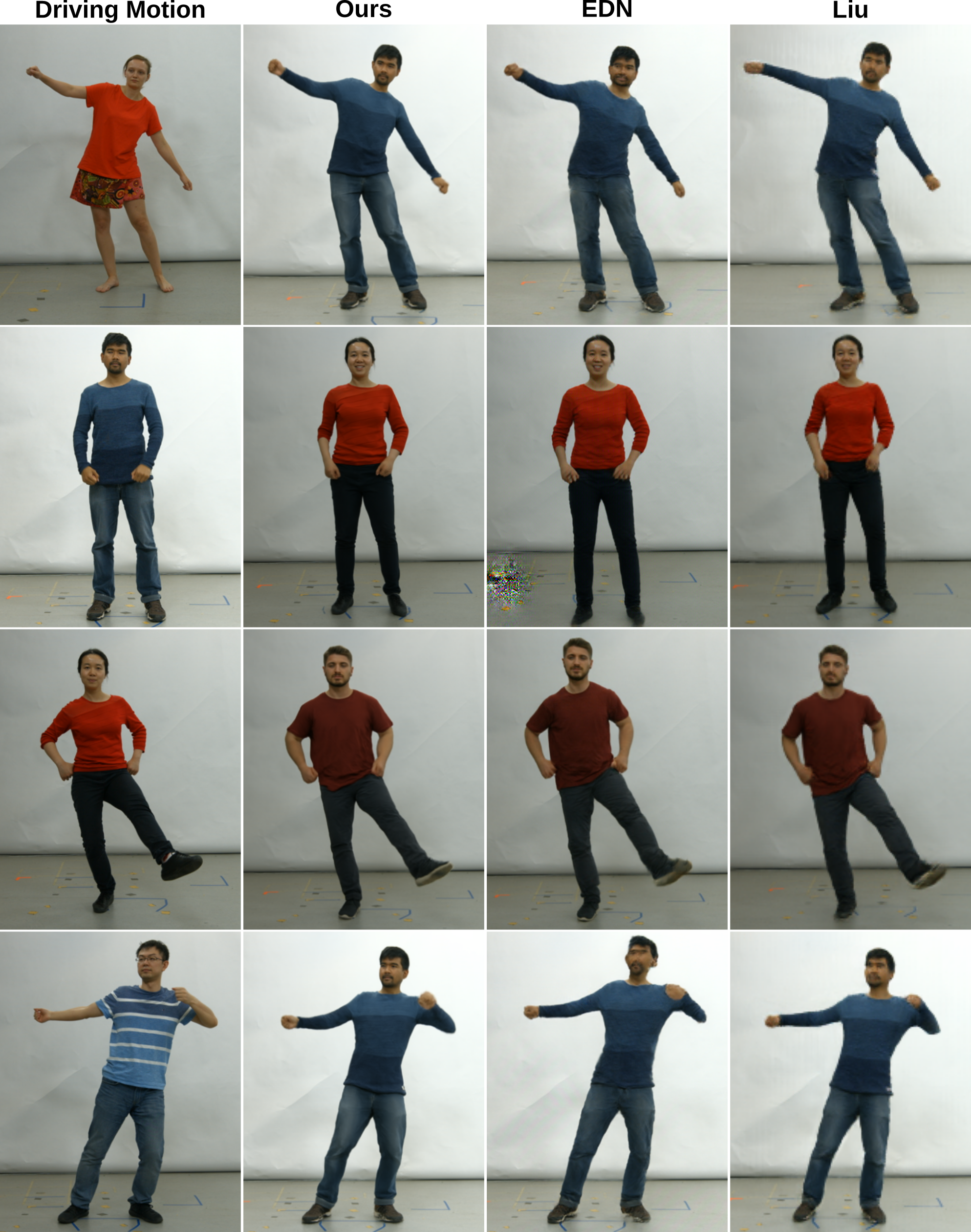}
    \vspace{-15pt}
    \caption{Exemplary frames from the video sequences used as  stimuli for our perceptual experiments. Top: \textit{Exp.~1}  conducted on our dataset. Bottom: \textit{Exp.~2} on the  dataset of Liu~\etal~\cite{liu2020neural}.} 
    \label{fig:comparisons} 
\end{figure} 
\paragraph{The First Experiment -- Our Dataset.}
As described in the main paper, we trained our method along with other three state-of-the-art reenacting techniques (EDN~\cite{Chan2019}, pix2pixHD~\cite{wang2018pix2pixHD}, and Recycle-GAN~\cite{bansal2018recycle}) on some sequences of our data set (see the top part of Fig.~\ref{fig:comparisons}).

\paragraph{The Second Experiment -- Liu~\etal's  Dataset.} 
We also evaluated our technique on the dataset of Liu~\etal~\cite{liu2020neural} (see the second bottom part of Fig.~\ref{fig:comparisons}). 
Here, we trained two of the compared methods (EDN~\cite{Chan2019} and ours) on this dataset.  Furthermore, we used the results of the reenactment approach of Liu~\etal that was provided by the authors.
Note we can not replicate their method on any other dataset as it requires a sophisticated capturing pipeline using a monocular structure-from-motion reconstruction of the target actor, including delicate pre-processing and mesh fitting. 
For fair comparison, all results were down-sampled to the resolution given by Liu~\etal (256x256).
Given the set of videos and the three tested methods (Liu~\cite{liu2020neural}, EDN~\cite{Chan2019} and ours), the total number of possible paired  comparisons reduces to twelve, making it well  suitable for a single participant to perform a complete test while  maintaining the necessary level of attention. 
%
\subsection{Experimental Procedure} 

\paragraph{The First Experiment.} 
The study was hosted online, and a total of $54$ subjects from various computer science backgrounds participated in the study. 
Before starting the experiment, participants were presented a short text describing the task and the procedure. 
The subjects were exposed to several video pairs that play side by side. 
Each video in a pair was produced by a different technique, and the order of pairing between methods, position on the screen and order of display of the pairs was fully randomized.
For each video pair, the subjects were asked to record their answers to two questions: Q1 (\textit{"Which video looks more realistic?"}) and Q2 (\textit{"Which video shows more natural motion and deformation of the clothes?"}).
The study included a total of twelve video pairs and took around ten minutes to complete. 

\paragraph{The Second Experiment.} 
The second experiment was conducted  \textit{in-situ}. 
A total of $16$ people (graphics and vision experts) participated (age range 22-40 years; four women). 
The mean time to complete the experiment was nine minutes. 
Before the experiment began, each participant was informed about the structure and flow of the experiment---but not of the research question behind it---and was given the option to ask any further questions. 
The participants performed the experiment one at a time. Before the experimenter left the room, the participant was asked to sit in a semi-dark room roughly 50 cm in front of a 24” LED monitor (with the resolution of $1920\times1080$). 
The respondents then were exposed to a screen describing in detail the instructions and were given another chance to ask questions.
The experiment was controlled by Psychophysics toolbox, version $3.0.15$ (PTB-3) \cite{psytb1,psytb2,psytb3}. 
At the start of each trial, participants were presented with two videos side by side. Participants were explicitly instructed to focus their attention on the garments and ignore any potential artifact not concerning the clothes. The participants were able to enter their answers by clicking on the desired video and also to replay each video individually as many times as desired; once a response was entered, the next trial started. 
Both the order of the stimuli, their pairing and their position on the screen (left \textit{vs} right) were fully randomized, with each participant receiving a different random order. Each participant reported normal or corrected-to-normal vision.

\subsection{Analysis}

To measure not only the performance of each method but the agreement between participants, we followed the linked-paired comparison design~\cite{david1963method}. Thus, we rank methods according to the number of times they are preferred over the other methods. The total number of votes a method received is displayed in 
 Table~\ref{tab:votesExpUpdated} for the first experiment (``\textit{Exp.~1}'') and Table~\ref{tab:Exp2} for the second experiment (``\textit{Exp.~2}''). 
%
\begin{table}[hbt!]
    \centering
    \setlength\tabcolsep{2 pt}
    {\footnotesize
    \begin{tabular}{|c|l|l|c|c|}
         \hline
         \textit{Exp.~1} & Method &Input & \#Votes & Ranking\\
         \hline
         \multirow{4}{*}{Q1}&\textbf{Ours} &Video (Pose) & 257&\textbf{1}\\
        &Recycle-GAN~\cite{bansal2018recycle} &Video (RGB) 
        & 194 &\textbf{2}\\
        &EDN~\cite{Chan2019} & Video (Pose)& 143&\textbf{3}\\
        &pix2pixHD~\cite{wang2018pix2pixHD} &Video (Pose)& 54&\textbf{4}\\
          \hline
         \multirow{4}{*}{Q2}&\textbf{Ours} &Video (Pose) & 243&\textbf{1}\\
        &Recycle-GAN~\cite{bansal2018recycle} &Video (RGB) 
        & 205 &\textbf{2}\\
        &EDN~\cite{Chan2019} & Video (Pose)& 135&\textbf{3}\\
        &pix2pixHD~\cite{wang2018pix2pixHD} &Video (Pose)& 65&\textbf{4}\\
         \hline
        \end{tabular}
    }
    \caption{\textit{Exp.~1}: Perceptual ranking of the compared methods for our online experiment with 54 participants with various backgrounds for each of the questions (Q1 and Q2). All the rankings are statistically significant.}
    \label{tab:votesExpUpdated}
\end{table}
\begin{table}[hbt!]
    \centering
    \setlength\tabcolsep{2 pt}
    {\footnotesize
    \begin{tabular}{|c|l|l|c|c|}
         \hline
         \textit{Exp.~2} &Method &Input & \#Votes & Ranking\\
         \hline
        \multirow{3}{*}{}&\textbf{Liu}~\cite{liu2020neural} & Textured mesh + Video (Pose) & 103& \textbf{1}\\
         &Ours & Video (Pose) & 76 & \textbf{2}\\
         &EDN~\cite{Chan2019} &Video (Pose) & 13 & \textbf{3}\\
         \hline
        \end{tabular}
    }
    \caption{\textit{Exp.~2}: Perceptual ranking of the compared methods on the in-situ user-study with $16$ CG/CV experts. The rankings are statistically significant. This table is already included in the main paper at the bottom of Table \ref{tab:votesExp} and is repeated  here for convenience. 
    }
    \label{tab:Exp2}
\end{table}

To analyze the true meaning of this ranking, we perform a significance test of the score differences. 
Towards that goal, we need to find a value $R'$ for which the variance-normalized range of scores within each group is lower or equal to that value. This means that we need to compute $R'$ such that $P[R \geq R'] \leq \alpha$, where $\alpha$ is the confidence level, which we set to $0.01$. Then, following the work of David~\cite{david1963method} we can derive $R'$ from
\begin{equation}
    P\bigg(W_{t,\alpha} \geq \frac{2R' - 0.5}  {\sqrt{mt}}\bigg),
\end{equation}
where $t$ is the number of methods to be compared, $m$ is the number of participants and $W_{t,\alpha}$ has been previously tabulated by Pearson and Hartley~\cite{Pearson1966Tables}. In our case, $W_{4,0.01} = 4.405$ for \textit{Exp.~1}  (Table~\ref{tab:votesExpUpdated}) and $W_{3,0.01}=4.125$ for \textit{Exp.~2} (Table~\ref{tab:Exp2}). This leads us to the values  $R'_{Exp1} = 32.62001$ for the first experiment and $R'_{Exp2} = 14.53942$ for the second experiment. 
Since all the differences between the ranked groups are bigger than the obtained $R'$, we can conclude that they are all statistically significant. Thus, the ranking creates four distinguishable groups in \textit{Exp.~1} for both Q1 (\textit{"Which video looks more realistic?"}) and Q2 (\textit{"Which video shows more natural motion and deformation of the clothes?"}). Furthermore, the ranking creates three distinguishable positions in \textit{Exp.~2} (\textit{"Which video displays more realistic clothes?  (movement/deformations/appearance)"}). 
%
\paragraph{Video Attributes.}
 Additionally, to gain more insight on the reasons to choose one result over another, in the experiment conducted \textit{in-situ} with CG/CV experts, the participants were occasionally asked to pick one or several items out of a proposed set of reasons for not choosing a result. This question appeared randomly with the probability of $1/3$, \textit{i.e.,} the frequency we found suitable in order to maintain the participant’s attention without making the test tedious. Table~\ref{tab:qs} shows the complete list of reasons and how often they were selected as a reason for rejecting a given result. As can be derived from the table, the most frequent reasons to discard a method were artifacts and implausible deformations of the clothing.
%
\begin{table}[hbt!]
    \centering
    \setlength\tabcolsep{2 pt}
    {\footnotesize
    \begin{tabular}{|l|r|}
        \hline
         \textbf{This is the video you did not choose in the last} & \% cases \\
         \textbf{comparison. Please specify which of the following} &  reason \\
         \textbf{bothers you in this video. You may check multiple options.} & selected\\
         \hline
         \hline
        Unrealistic wrinkles. & 37.5\%\\
        Unrealistic clothes' texture. & 23.44\%\\
        Implausible deformations of the clothing. & \textbf{59.63\%}\\
        Temporal inconsistencies in the clothes' movement. & 20.31\%\\
        Other artifacts in clothing. & \textbf{45.31\%}\\
        The other result was simply more appealing. & 10.94\%\\
        Other. & 3.21\%\\
      \hline
     \end{tabular}
     }
     \vspace{2pt} 
    \caption{Questionnaire displayed after rejecting a result. The participant was able to choose as many answers as desired. The second column shows the frequency (\%) for reporting one type of artifact when the questionnaire appeared after rejecting a result (four times per participant). These numbers  indicate the most frequent reasons to discard a  method, and thus, the most important features for  participants.} 
    \label{tab:qs} 
\end{table} 

%
\section{Training Details}
Our framework $\boldsymbol{G}_\theta$ consist of four trainable components, namely $\boldsymbol{G_{ref}}$, $\boldsymbol{G_{app}}$, $\boldsymbol{G_{str}}$, $\boldsymbol{G_{shp}}$.
For every network, we adapt the local Pix2PixHD~\cite{wang2018pix2pixHD} generator architecture, resulting in memory consumption and training times of ${\sim}3.5 \times$ the reported values for the original implementation ($\boldsymbol{G_{ref}}$ uses a reduced amount of intermediate blocks due to the simplicity of the learned mapping).
However, to enable training on current graphics hardware, every instance is currently trained individually using annotations and losses described in the main document, which allows for stepwise processing on consumer-grade devices, requiring ${\sim}5$GB VRAM and ${\sim}280$ milliseconds per frame. 
To achieve temporal smoothness in our garment conditioning modules---and, thus, temporally consistent renderings of body parts and clothing---we condition the shape and structure predictions on the  previous outputs of the networks. 
Intuitively, this information is crucial to estimate the non-rigid deformations within clothing, as they do not only depend on forces extracted from the change of pose, but also the current state of the dynamic system.
Thus, instead of computing two consecutive frames and applying a temporal discriminator, we process the videos in an entirely sequential manner, which results in temporally more stable results, as shown in our supplemental video.
Still, we cut the gradient to previous outputs, as we find that truncated temporal backpropagation significantly increases memory consumption and training times, without contributing much to the final quality.
To further stabilize initial training, we feedback pseudo-ground-truth annotations during the first epoch, which encourages the network to expect reliable estimates of previous clothing deformations.
We further synthesize the first frame of every video based on a black image (as no predecessor is available)  and iteratively execute our networks on the initial input pose until the output converges towards a reasonable estimate (in practice, this takes around 20-30 iterations).


\end{document}